\documentclass[sigconf]{acmart}
\usepackage{graphicx}
\usepackage{caption}
\usepackage{marvosym}
\usepackage{subcaption}
\usepackage{colortbl}
\usepackage{hyperref}
\hypersetup{hidelinks,colorlinks=true,allcolors=black,pdfstartview=Fit,breaklinks=true}
\AtBeginDocument{%
  }

\copyrightyear{2024}
\acmYear{2024}
\setcopyright{acmlicensed}

\acmConference[MM '24 ]{Proceedings of the 32nd ACM International Conference on Multimedia}{October 28--November 1, 2024}{Melbourne, VIC, Australia.} 
\acmBooktitle{Proceedings of the 32nd ACM International Conference on Multimedia (MM '24), October 28--November 1, 2024, Melbourne, VIC, Australia} 
\acmDOI{10.1145/3664647.3681085 }
\acmISBN{979-8-4007-0686-8/24/10 }
\begin{document}


\title{MMDRFuse: Distilled Mini-Model with Dynamic Refresh for Multi-Modality Image Fusion}

\author{Yanglin Deng}
\email{yanglin_deng@163.com}
\orcid{0009-0005-7022-2869}
\affiliation{%
  \institution{Jiangnan University}
  \city{Wuxi}
  \country{China}
}

\author{Tianyang Xu\textsuperscript{\Letter}}
\email{tianyang.xu@jiangnan.edu.cn}
\orcid{0000-0002-9015-3128 }
\affiliation{%
  \institution{Jiangnan University}
\city{Wuxi}
  \country{China}
}

\author{Chunyang Cheng}
  \email{chunyang_cheng@163.com}
  \orcid{0000-0003-4603-3505  }
\affiliation{%
  \institution{Jiangnan University}
  \city{Wuxi}
  \country{China}
}

\author{Xiao-Jun Wu}
\email{wu_xiaojun@jiangnan.edu.cn}
  \orcid{0000-0002-0310-5778}
\affiliation{%
 \institution{Jiangnan University}
 \city{Wuxi}
 \country{China}
 }

\author{Josef Kittler}
\email{j.kittler@surrey.ac.uk}
  \orcid{0000-0002-8110-9205 }
\affiliation{%
  \institution{University of Surrey}
  \city{Guildford}
  \country{United Kingdom}
  }

\renewcommand{\shortauthors}{Yanglin Deng, Tianyang Xu, Chunyang Cheng, Xiao-Jun Wu, \& Josef Kittler}

\begin{abstract}
In recent years, Multi-Modality Image Fusion (MMIF) has been applied to many fields, which has attracted many scholars to endeavour to improve the fusion performance. However, the prevailing focus has predominantly been on the architecture design, rather than the training strategies. As a low-level vision task, image fusion is supposed to quickly deliver output images for observation and supporting downstream tasks. Thus, superfluous computational and storage overheads should be avoided. In this work, a lightweight Distilled Mini-Model with a Dynamic Refresh strategy (MMDRFuse) is proposed to achieve this objective. To pursue model parsimony, an extremely small convolutional network with a total of 113 trainable parameters (0.44 KB) is obtained by three carefully designed supervisions. First, digestible distillation is constructed by emphasising external spatial feature consistency, delivering soft supervision with balanced details and saliency for the target network. Second, we develop a comprehensive loss to balance the pixel, gradient, and perception clues from the source images. Third, an innovative dynamic refresh training strategy is used to collaborate history parameters and current supervision during training, together with an adaptive adjust function to optimise the fusion network. Extensive experiments on several public datasets demonstrate that our method exhibits promising advantages in terms of model efficiency and complexity, with superior performance in multiple image fusion tasks and downstream pedestrian detection application. The code of this work is publicly available at \href{https://github.com/yanglinDeng/MMDRFuse}{\textcolor{blue}{https://github.com/yanglinDeng/MMDRFuse}}.
\end{abstract}

\ccsdesc[500]{Information systems~Multimedia content creation}

\keywords{multi-modality image fusion, feature-level distillation, dynamic refresh, end-to-end training}
\begin{teaserfigure}
\captionsetup{skip=2pt}
  \includegraphics[width=\textwidth]{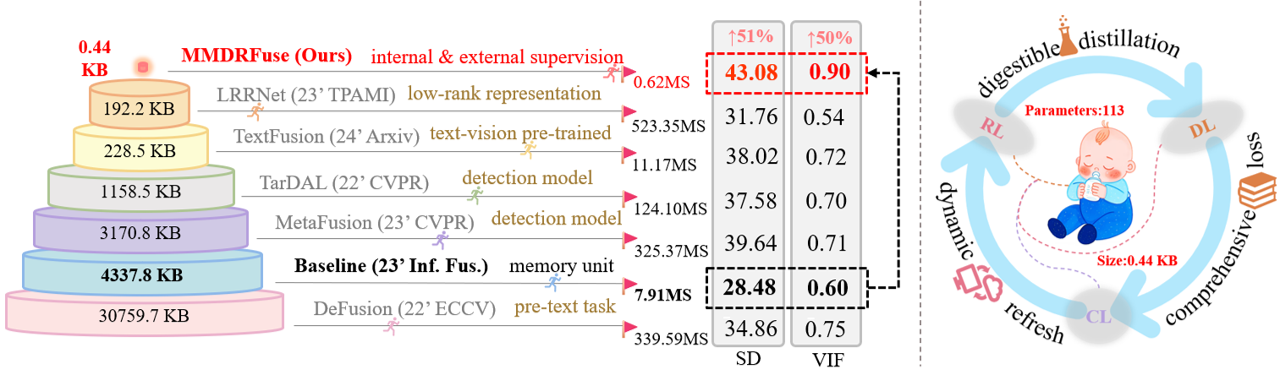}
  \caption{Left: Comparison of our Mini Model (MMDRFuse) with other advanced image fusion solutions in terms of model efficiency, including model size (KB), average running time (MS), and two image quality assessments SD and VIF. 
Right: Supervision designs of our MMDRFuse, where digestible distillation refers to delivering external soft supervision, dynamic refresh emphasises the historical impact of network parameters during training, and comprehensive loss serves as an internal supervision that absorbs nutrients from source images.}
\Description{The supervision strategy and performance of MMDRFuse}
  \label{fig:motivation}
\end{teaserfigure}

\maketitle

\section{Introduction}
Images captured by different physical sensors under diverse conditions contain unique attributes, which challenge the design of a general image processing system.
Drawing on this, to unify the visual pixel distribution at the image level, image fusion provides a solution to combine source images into a single and comprehensive output image~\cite{overview}. 
Considering the input configurations, image fusion tasks are ranging from digital image fusion \cite{1998Progress, 2021Deep}, multi-modality image fusion (MMIF) \cite{U2Fusion}, to remote sensing image fusion~\cite{2021SIPSA,2021yaogan}.
In particular, MMIF~\cite{CDDFuse, DDFM, TarDAL} has attracted wide attention in recent decades, which includes Infrared and Visible Image Fusion (IVIF) \cite{2020RXDNFuse, SeAFusion}, and Medical Image Fusion (MIF) \cite{2018Medical}.

In terms of the IVIF task, the target sources are visible and infrared images \cite{CDDFuse}.
Specifically, the visible modality, captured by optical devices such as digital cameras, is expert at preserving textural details and colours, while it is susceptible to illumination conditions.
On the other hand, the infrared modality, derived from infrared imaging devices, captures the inherent heat radiation emitted by living beings or powered objects.
This modality is obtained based on the variation in radiation intensity between the target and its surroundings, lacking supportive colours and textures as contained in the visible spectrum.
The IVIF task aims to generate an informative image that leverages the strengths of both, enhancing the overall visual quality and serving various following processing demands \cite{overview, DDFM}.
Specifically, the fused images can be applied to several downstream tasks, \textit{e.g.}, semantic segmentation \cite{2021Searching, 2022Defensive}, object tracking \cite{Object, 2020SiamRPN}, object detection \cite{object_detection, object_detection2,target_detection}, and saliency detection \cite{salience}.
As for the MIF task, by combing images obtained through different medical imaging devices (CT \cite{ct}, MRI, PET, SPETC \textit{e.t.c}.), MIF can provide a clearer view of both the structure and functional information of the human tissues and organs, supporting more precious disease diagnosis~\cite{2018Medical, DDFM}. 

After decades of study, the performance of IVIF has been greatly advanced.
In contrast to traditional approaches that rely on multi-scale decomposition (MSD) \cite{MSD}, sparse representation (SR) \cite{SR}, or low-rank representation (LRR) \cite{LRR}, to extract handcrafted features,
more robust representations can be obtained by deep solutions, such as convolutional neural networks (CNNs), Transformers, and their hybrid versions~\cite{overview}.
To pursue better visual effects, various functional blocks have been developed, \textit{ e.g.}, aggregated residual dense block \cite{2020RXDNFuse} and gradient residual dense block \cite{SeAFusion}. 
To facilitate smooth training, residual connections \cite{2018DenseFuse, RFNnest} and skip connections~\cite{MUFusion} are adopted to prevent gradient vanishing or distortion in the generated images.
Besides modifying network modules, more efforts have been explored to formulate fusion as a generation task.
For instance, \cite{TarDAL,DDcGAN} utilise a generator to obtain the fused image, accompanied by two discriminators to guarantee its fidelity.
Essentially, the generator contains multiple CNN layers, exhibiting the potential of crash and control issues in the adversarial training scheme. 
Such computation burden is also suffered by the diffusion solution \cite{DDFM}, which obtains stable and controllable high-quality fused images without discriminator but requires more computation resources than previous approaches. 

Despite the performance promotion achieved by the above attempts, all the involved network designs suffer from excessive complexity and redundancy.
Therefore, striking a balance between model performance and resource requirements is a pressing issue that needs to be addressed.  
Motivated by knowledge distillation \cite{firstdistill, attentiondistill}, it is straightforward to compress the teacher parameters into a relatively smaller student model.
We propose establishing a strong teacher model with powerful feature extraction and reconstruction ability.
To deliver effective supervision, as shown in Figure~\ref{fig:motivation}, we adopt three dedicated designs to integrate supervision signals from the teacher, source images, and history records.
Specifically, a digestible distillation strategy is proposed to relax the strict consistency constraints between the student network and the teacher network on the intermediate feature dimension. 
To exploit the source input images, intensity details, edge gradients, and perception semantics are comprehensively reflected by our loss function.
Furthermore, to reinforce the cues embedded in the historical parameters of the training process, we propose the dynamic refresh strategy.  
A set of dual evaluation metrics are devised to distinguish whether the current cues are useful or not. 
The useful record demonstrates that our training trajectory is correct at present and can be refreshed as internal supervision signals, while the inadequate one can be optimized by the above signals.



In this paper, we design a \textit{Distilled Mini-Model with Dynamic Refresh for Multi-Modality Image Fusion} (MMDRFuse). 
As shown in Figure \ref{fig:motivation}, our MMDRFuse achieves promising fusion performance, while it is significantly smaller than other SOTA models, with only 113 trainable parameters (0.44 KB).
The contributions of our work can be summarised as follows:
\begin{itemize}
    \item  An end-to-end fusion model with a total of 113 trainable parameters and 0.44KB size, which can efficiently facilitate fusion and support downstream tasks. 
    \item A digestible distillation strategy to relax the feature-level consistency, softening the supervision from the teacher model.
    \item A comprehensive loss function to preserve intrinsic clues from source inputs, collaborating pixels, gradients, and perceptions.
    \item  A dynamic refresh strategy to effectively manage the historical states of parameters during training, endowed with a dual evaluation metric system to adaptively refine supervision signals towards the correct direction. 
    \item Promising performance in terms of fusion quality and efficiency against SOTA solutions. The obtained 113 parameters can even support advanced downstream pedestrian detection performacne.
\end{itemize}
\section{Related Work}
\subsection{Advanced MMIF Formulations}
Recent studies in MMIF can be broadly categorised into Auto-Encoder (AE) paradigm \cite{2018DenseFuse, RFNnest, CDDFuse, textfusion, 2022DeFusion}, GAN-based paradigm \cite{TarDAL, FusionGAN}, Transformer-based paradigm \cite{CDDFuse, SwinFusion}, downstream task-driven paradigm \cite{TarDAL, MetaFusion, SeAFusion}, and text-driven paradigm \cite{textfusion}.
Among the involved techniques, AE exploits an encoder to extract features from source images and a decoder to obtain fused images by reconstructing the latent features. 
To get rid of the limitations of handcrafted fusion rules, \cite{RFNnest} devises a residual fusion network to smooth the training stage. 
To alleviate computation, \cite{CDDFuse} uses Lite-Transformer to fuse base features and detail features come from different modalities. 
Besides network design, \cite{MUFusion} directly conducts an end-to-end fusion network with a memory unit to allocate effective supervision signals. 
\begin{figure*}[t]
  \centering
  \includegraphics[width=\linewidth]{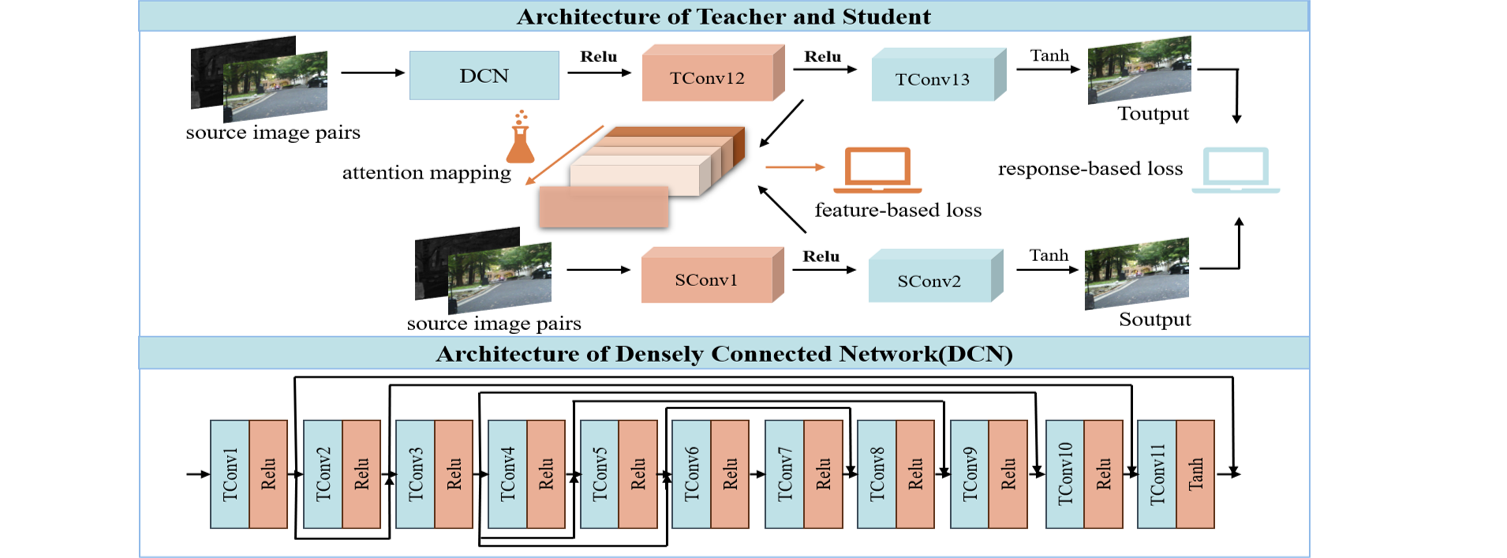}
  \caption{Illustration of our distillation process. TConv1, TConv2, ... , TConv13 represent the convolutional layers in the teacher network. SConv1, SConv2 represent the convolutional layers in the student network.  TOutput and SOutput denote their outputs respectively.}
\Description{Architecture of teacher and student model with digestible distillation mechanism.}
\label{distillation1}
\vspace{-3mm}
\end{figure*}
To better support downstream tasks, \cite{TarDAL} proposes a bilevel optimisation formulation based on GANs and Object Detection (OD) network, forming a cooperative training scheme to yield optimal network parameters with fast inference for both tasks. 
Besides, \cite{SeAFusion} introduces a real-time segmentation model to grind the semantic information for the fused images.  
Although linking the downstream tasks with image fusion can obtain guided semantics, such a solution sacrifices too much in computation, running time, and storage space. 
To break away from existing modelling techniques, we investigate the possibility of designing internal and external supervision signals to serve an extremely tiny model.
In particular, the internal and external supervision denote the historical state of the model itself and the guidance of a powerful teacher model, respectively.
Such a combination is typically controlled by a dynamic refresh strategy to harmonise the training of our 0.44KB mini model.

\subsection{Distillation Techniques}
Knowledge Distillation is introduced by \cite{firstdistill} to tackle the cumbersome of deploying computationally expensive models.
According to the locations of processing knowledge, it can be categorised into three branches, \textit{i.e.}, response-level \cite{firstdistill, 2022Decoupled}, feature-level \cite{2015FitNets, attentiondistill, reviewdistill, 2021Rethinking, 2020Channel} and relation-level \cite{relation1,2019Relational} settings.
Typically, \cite{firstdistill} distils knowledge by aligning the student's output with the teacher's output while \cite{2015FitNets} additionally focuses on the consistency of intermediate features between teacher and student. 
Following this setting, advanced studies are dedicated to paying attention to feature-based distillation.
In terms of formulating feature consistency, \cite{attentiondistill} proves that activation-based attention transfer and gradient-based spatial attention transfer are more effective than full-activation transfer.
Considering feature diversity, \cite{reviewdistill} proposes to gradually learn the low-level feature maps after the high-level consistency is obtained.
However, this knowledge delivery solution still costs a lot of resources during the training process. 

Given that the structure of our student network is too simple (113 parameters) to directly learn from a powerful teacher, similar to feeding a baby, how to transfer digestible supervision from the teacher to the student poses significant challenges. 
Drawing on this, we feed the student with more finely processed and swallowable knowledge by combining feature-based and response-based distillation within isomorphic transformation modules. 

\section{Approach: MMDRFuse}

\subsection{Digestible Distillation}
In this subsection, we introduce our solution to absorb digestible supervision from the teacher's guidance.
In essence, the ultimate goal of our MMDRFuse is to obtain a mini model with satisfactory fusion performance and downstream supportive power. 
It is impossible to directly train the mini model itself, as it is greatly limited by its parameter volume. 
Accordingly, we first train a teacher model, which is expected to exhibit considerable feature extraction and reconstruction ability to feed our mini model.

Existing studies unveil that aligning knowledge at the output end is too violent for the student model.
In particular, different from high-level visual tasks, such as classification and detection, the output of a fusion system is low-level pixels, which challenges the strictness of teacher guidance.
Therefore, we propose to add a buffer before the final output to transfer the supervision that can be digested better. 
At the same time, feature maps of the middle layers are supposed to be consistently distributed as the teacher model \cite{attentiondistill}, which can help to produce the expected output.

We adopt a relatively complex teacher network to obtain a robust fusion model.
This network incorporates a densely connected network (DCN) responsible for extracting deep image features comprising 64 channels.
Additionally, it integrates two convolutional layers to generate four-channel feature maps and the single-channel output, respectively.
By contrast, our student model only comprises two convolutional layers.
The first one is used to extract feature maps with four channels and the latter is designed as a decoder to produce the fused image.
We implement feature-based and response-based distillation with the last two layers between the teacher and student, respectively. 
Figure \ref{distillation1} illustrates the network architecture and distillation mechanism.

To ensure that the student outputs closely resemble those of the teacher network, we encourage consistency between their outputs.
Specifically, the feature-based distillation and the response-based distillation can be formulated as:
\begin{equation}
    L_{distill} =  \frac{1}{2}\sum^2_{i=1}\Arrowvert \frac{vec(F(T_{i}))}{\Arrowvert vec(F(T_{i})) \Arrowvert_2}-\frac{vec(F(S_{i}))}{\Arrowvert vec(F(S_{i})) \Arrowvert_2} \Arrowvert_2 ,
\end{equation}
where $vec(\cdot)$ denotes the vectorisation operation.
$F(X)=\sum_{i=1}^{C} x_{i}$ is a spatial mapping function used to conduct attention mapping across channel dimensions, where $x_i$ is the $i-th$ channel of $X$.
$F(X)$ comprehensively considers all channels from $X$ by distributing average weights to each spatial region.
On the other hand, for the feature-based distillation, $S_{1}$$\in$ $R^{ H\times W\times 4}$ and $T_{1}$ $\in$ $R^{ H\times W\times 4}$ are the extracted features of the teacher network and student network, respectively.
For the response-based distillation, $S_{2}$ $\in$ $R^{ H\times W\times 1}$ and  $T_{2}$ $\in$ $R^{ H\times W\times 1}$ are the output images of these two networks.
As these output feature maps and images share an identical number of channels, the teacher network can effectively deliver knowledge to the student module at the same semantic level.

\subsection{Comprehensive Loss}
\begin{figure}[t]
\centering
\includegraphics[width=\linewidth]{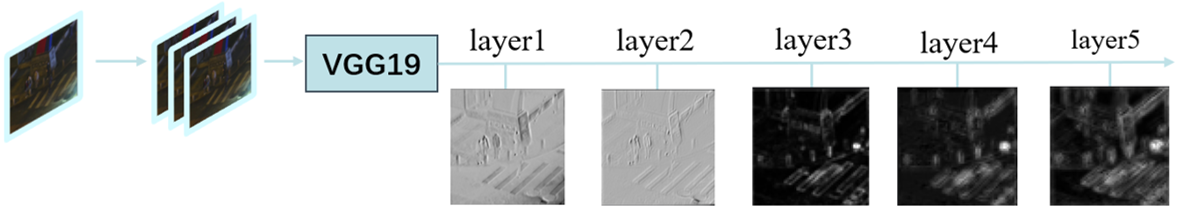}
\caption{Illustration of the feature maps used to reflect perception degrees. From left to right, it represents the source image, duplicated source image, and five feature maps extracted by VGG-19, respectively.
}
 \Description{Multi-layer extracted feature maps by VGG-19.}
\label{vgg}

\vspace{-3mm}

\end{figure}
In this section, we introduce our solution to absorb valid supervision from the source inputs.
Firstly, we expect the fused image to be similar to source images at the pixel level.
Specifically, for the multi-modality image fusion task, visible or MRI images usually contain texture details, while infrared or PET images tend to present significant thermal and functional information.
Thus, an intensity loss function~\cite{SeAFusion} is designed to generate fused images that share the same pixel distribution with source inputs, \textit{i.e.}
\begin{equation}\label{int}
    L_{\textrm{int}} =  \frac{1}{HW}\Arrowvert O- max{(I_{\textrm{ir}},I_{\textrm{vis}})} \Arrowvert_1 ,
\end{equation}
where $I_{\textrm{ir}}$ and $I_{\textrm{vis}}$ denote the source infrared and visible images, $O$ represents the output of the fusion network.
Notably, if source images are under normal light conditions, Eqn.~\eqref{int} can help to maintain detailed targets and clear structure.
However, if the light condition in texture areas is darker than the infrared image, the intensity is dominated by the infrared signals. 
Thus, it seems not enough to rely solely on the intensity loss function.
Therefore, we use a maximum gradient loss to help transfer gradient information mainly from the visible image to the target, it is defined as:
\begin{equation}
    L_{grad} =  \frac{1}{HW} \Arrowvert \bigtriangledown O-\bigtriangledown max{(I_{\textrm{ir}}, I_{\textrm{vis}})} \Arrowvert_F^{2} ,
\end{equation}
where $\Arrowvert \cdot \Arrowvert_{F}$ represents the Frobenius norm,$\bigtriangledown$ represents gradient operator.
Furthermore, to better fuse fine-grained features from source images, we employ a normalised VGG-19 network \cite{2020RXDNFuse} to extract hierarchical features of input and output images, formulating a feature-based perception loss.
As depicted in Figure \ref{vgg}, the feature maps in the shallow layers exhibit clearer edge texture details and gradient information, while the feature maps in deep layers become blurred, encompassing higher-level semantic information, abstract contexts, and salient structure information.
This information is a significant factor regarding the fusion performance~\cite{MUFusion}.
Hence, the perception loss is formulated as:
\begin{equation}
    L_{\textrm{percep}} = \sum_{i=1}^{5} \frac{ \sum_{j=1}^{D_{i}} \Arrowvert  \phi_{i,j}(O)- max{(\phi_{i,j}(I_{\textrm{ir}}),\phi_{i,j}(I_{\textrm{vis}}))} \Arrowvert_F^{2}} {5 \cdot H_{i} \cdot W_{i} \cdot D_{i}} ,
\end{equation}
where $D_i$ represents the number of channels of extracted feature map $i$, $H_{i}$ and $W_{i}$ represent the height and width of feature map $i$, $\phi_{i,j}$ represents the $j-th$ channel of the $i-th$ feature map. 
Finally, the above components collectively constitute the comprehensive loss function:
\begin{equation}
    L_{comp} =   \gamma L_{int}+ \delta L_{grad}+L_{percep}.
\end{equation}
By adding the maximum gradient loss and maximum perception loss, our fusion network not only focuses on regions with high brightness but also pays attention to texture information in dark areas, which cannot be achieved solely by intensity loss. 
At the same time, without intensity loss, we can not retain salient targets only by gradient loss and perception loss.
Hence, the combination of these loss functions can complement each other's limitations while keeping their respective strengths.
\subsection{Dynamic Refresh Strategy}
\begin{figure}[t]
\centering
\includegraphics[clip,width=0.9\linewidth]{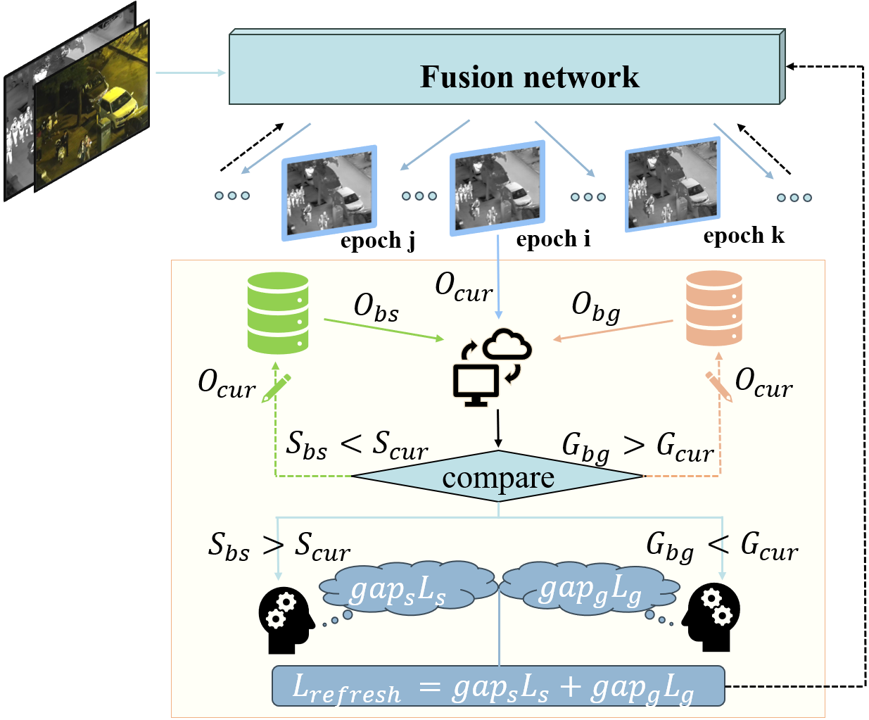}

\caption{Workflow of the proposed dynamic refresh strategy. The green and orange databases denote the best historical fusion outputs measured by SSIM and GMSD, respectively. The centre represents the process of calculating $S_{bs}$ and $S_{cur}$, $G_{bg}$ and $G_{cur}$. The brains symbolize the process of calculating two loss items. The black dashed line denotes the process of utilising $L_{refresh}$ to supervise the training process. 
}
\Description{Workflow of the dynamic refresh strategy.}
\label{refresh}

\vspace{-3mm}
\end{figure}
Traditional methods often directly discard the states of parameters from intermediate iterations during the training process, which, we argue, can be properly utilised to serve as supervision signals. 
However, useful information can not be guaranteed in the intermediate iterations, sometimes including noises, artefacts and so on. 
Hence, we need to identify the effectiveness of them. 
We devise a dual evaluation system to identify and further utilise the advantageous parameters. 
We call this process as dynamic refresh, Figure \ref{refresh} illustrates the basic process. 
In particular, dynamic refresh is applied throughout the entire training process, and we adopt two metrics to measure whether it is helpful to fusion.
Image quality assessment structural similarity (SSIM) \cite{ssim} and Gradient Magnitude Similarity Deviation (GMSD) \cite{gmsd} are involved as the metrics.
SSIM measures the previous iterations from the perspective of structure, illuminance, and contrast, while GMSD highlights texture details. 
Accordingly, we measure every output during the training process and keep the two best outputs, which correspond to the green database and the orange database in Figure \ref{refresh}, respectively.
The evaluation process can be described as follows:
\begin{equation}\left\{
\begin{aligned}
    S_{bs} = &  SSIM(O_{bs},I_{ir})+ SSIM(O_{bs},I_{vis})\\
    S_{cur} = &  SSIM(O_{cur},I_{ir})+ SSIM(O_{cur},I_{vis})
\end{aligned}\right.,
\end{equation}
\begin{equation}\left\{
\begin{aligned}
    G_{bg} = & GMSD(O_{bg},I_{ir})+GMSD(O_{bg},I_{vis})\\
    G_{cur} = & GMSD(O_{cur},I_{ir})+GMSD(O_{cur},I_{vis})
\end{aligned}\right.,
\end{equation}
where $O_{cur}$ symbols the output of the current epoch, $O_{bs}$ symbols the historical output with the best SSIM value, $O_{bg}$ symbols the historical output with the best GMSD value. 
$S_{bs}$,$S_{cur}$ represent the SSIM value of the best history output and the current output, respectively. 
$G_{bg}$,$G_{cur}$ represent the GMSD value of the best history output and the current output, respectively.

By calculating SSIM and GMSD values of the current output and the two best historic outputs, once the value of the current output is better than the recorded outputs, they will be replaced by the current output, which realises the dynamic update. 
The refreshing process can be described as follows:
\begin{equation}
\begin{aligned}
    O_{\text{bs}} =
\begin{cases}
  O_{\text{bs}}, &  S_{\text{bs}} > S_{\text{cur}} \\
  O_{\text{cur}}, &  S_{\text{bs}} < S_{\text{cur}}
\end{cases}
\end{aligned},
\quad
\begin{aligned}
 O_{\text{bg}} =
\begin{cases}
  O_{\text{bg}}, &  G_{\text{bg}} < G_{\text{cur}} \\
  O_{\text{cur}}, &  G_{\text{bg}} > G_{\text{cur}}
\end{cases}.
\end{aligned}
\end{equation}

On the contrary, the current output can learn from historic outputs by following refresh loss:
\begin{equation}
\begin{aligned}
    L_{s} = \sum_{i=4}^{5}\frac{ \sum_{j=1}^{D_{i}} \Arrowvert  \phi_{i,j}(O_{cur})-\phi_{i,j}(O_{bs}) \Arrowvert_F^{2}}{2 \cdot H_{i} \cdot W_{i} \cdot  D_{i} }+ \frac{\Arrowvert O_{cur}- O_{bs} \Arrowvert_1}{HW}.
\end{aligned}
\end{equation}
\begin{equation}
{\footnotesize
\begin{aligned}
    L_{g} =\sum_{i=1}^{3} \frac{ \sum_{j=1}^{D_{i}} \Arrowvert  \phi_{i,j}(O_{cur})-\phi_{i,j}(O_{bg}) \Arrowvert_F^{2}}{3 \cdot H_{i} \cdot W_{i} \cdot  D_{i}}+\frac{ \Arrowvert \bigtriangledown O_{cur} - \bigtriangledown O_{bg} \Arrowvert_F^{2}}{HW}.\end{aligned}}
\end{equation}
Motivated by the phenomenon presented in Figure~\ref{vgg}, we combine perception loss of the last two feature maps before the max-pooling operation with intensity loss function, to collectively serve as supervision signals by the item of $L_s$. 
Similarly, we combine perception loss of the first three feature maps before the max-pooling operation with maximum gradient loss function, to collectively serve as supervision signals by the item of $L_g$.
$L_s$ can help to fuse context and spatial structure, brightness, and contrast information from $O_{bs}$, while $L_g$ can help to fuse texture detail and gradient information from $O_{bg}$. 
Furthermore, we measure the gap between $S_{bs}$ and $S_{cur}$, as well as $G_{bg}$ and $G_{cur}$, to serve as two self-adaptive coefficients:
\begin{equation}\left\{
\begin{aligned}
    gap_{s} =& S_{bs}-S_{cur}  \\
    gap_{g} =& G_{cur}-G_{bg}  
\end{aligned}\right..
\end{equation}
Hence, our dynamic refresh loss can be described as follows:
\begin{equation}
    L_{refresh} =  gap_{s} L_{s}+ gap_{g} L_{g}.
\end{equation}
Finally, we use comprehensive loss and refresh loss to collectively train our teacher model and further add the distillation loss to train our student model. 
The total loss function can be described as follows:
\begin{equation}
    L_{total} =  \theta L_{distill} +  \lambda  L_{comp}+ L_{refresh}.
\end{equation}

\section{Evaluation}
\subsection{Setup}
\subsubsection{Datasets and Metrics}
We verify MMDRFuse on three types of fusion tasks: Infrared and Visible Image Fusion (IVIF), Medical Image Fusion (MIF), and Pedestrian Detection (PD). 
We select six SOTA methods to compare with our method, including TextFusion \cite{textfusion}, MUFusion \cite{MUFusion}, LRRNet \cite{li2023lrrnet}, MetaFusion \cite{MetaFusion}, DeFusion \cite{2022DeFusion}, and TarDAL \cite{TarDAL}.
To better prove the generalisation ability of our design, we only train our model (both teacher and student) on the IVIF task, and directly apply the trained model to other tasks.
The training dataset is LLVIP \cite{2021LLVIP}.
We select 200 image pairs from LLVIP, which are further randomly cropped into 128$\times$128 image patches, ending with 16000 image pairs. 
For the MMIF task, we conduct test experiments on three datasets: MSRS~\cite{MSRS}, LLVIP, and RoadScene \cite{2020FusionDN}. 
The image numbers of them are 361, 250, and 50, respectively. 
For the MIF task, we directly test the model on MRI-SPECT and MRI-PET, which include 73 and 42 image pairs, respectively. 

For the PD task, we train six SOTA methods and our method on 2000 image pairs from the LLVIP dataset, and then test the detection performance on the 250 test image pairs following the LLVIP protocol. 
We perform PD experiments by using the Yolov5\cite{yolov5} as a detector to evaluate the pedestrian detection performance with the value of mAP@.5:.95, which provides a more comprehensive evaluation by calculating the average precision according to the IOU threshold from 0.5 to 0.95, taking into account the performance of the model throughout the entire retrieval process. 
The training epoch and batch size for PD evaluation are set as 3 and 16, respectively, with the SGD optimiser.

Besides, we adopt six metrics to fairly evaluate fusion performance, including standard deviation (SD) \cite{SD}, Sum of Correlation Differences (SCD)~\cite{SCD}, visual information fidelity (VIF) \cite{VIF}, structural similarity index measure (SSIM) \cite{ssim}, $\mathrm{Q^{AB/F}}$ \cite{QABF}, and Correlation Coefficient (CC) \cite{SCD}. 
SD is a statistical metric, that measures an image from its brightness and contrast, but sometimes it can be disturbed by noise. 
SCD characterises the quality of the fusion algorithm by measuring the differences between the fused image and the source images.
VIF not only considers the pixel-level similarity with source images, but also cares about human visual perception. 
SSIM measures fused images from structure, illuminance, and contrast. 
$Q^{AB/F}$ not only focuses on the visual quality of the image but also takes into account the preservation of the image content and structure.
CC is a metric that measures the degree of linear correlation between the fused image and the source image.

\subsubsection{Implement Details}
Our experiments are conducted on a Linux server with an NVIDIA GeForce RTX 3090 GPU. 
We first train the teacher model with 8 epochs and the number of batch size is 36.
Then we adopt the trained teacher model to collectively train the student model with 10 epochs and the batch size is 26. 
For hyperparameters setting, we set $\gamma=2$, $\delta=0.1$, $\theta=0$, and $\lambda=1$ for teacher model, and $\gamma=0.5$, $\delta=0.05$, $\theta=5$, and $\lambda=1$ for student model, respectively. 
In terms of the optimiser, we utilise the Adam optimizer with an initial learning rate equal to $10^{-4}$.
\subsection{Ablation Studies}
In this part, we aim to demonstrate the contribution of each component of our proposed mini-model.
Notably, our comprehensive loss function contains three distinct items, and the efficacy of each can be individually assessed by observing the performance degradation when they are omitted. 
This approach can also be applied to evaluate the functionality of the dynamic refresh strategy and digestible distillation.
\subsubsection{Comprehensive Loss Function and Dynamic Refresh Strategy}
As we mentioned in the comprehensive loss function part, intensity loss is mainly used to retain the pixel distribution, salient targets, and general structural information. 
The maximum gradient item and maximum perception item are further utilised to retain texture, gradient, and high-level semantic information. 
Only through their joint supervision can the information from the source images be fully integrated and utilized. 
This phenomenon can be observed from the first three rows of Table \ref{tab:ablation}. 
The omission of any component in the comprehensive loss leads to a decline in the overall performance, especially when the perceptual loss is missing.


Our dynamic refresh strategy is mainly adopted to sufficiently explore the vital information contained in the intermediate training stage. 
With the combined effect of comprehensive loss functions, the extraction power becomes increasingly strong, fully absorbing the information contained in not only the source images but also the historical recorded outputs.  
From the first four rows of Table \ref{tab:ablation}, we can observe that, even with the help of comprehensive loss and digestible distillation, our mini-model can not reach promising performance without involving the dynamic refresh strategy.

\begin{table}
  \caption{Quantitative ablation study for each component of our mini-model on MSRS. The boldface shows the best value.}
  \captionsetup{skip=2pt}
  \label{tab:ablation}
  \begin{tabular}{ccccc}
    \toprule
    Configurations&SD&VIF & $\mathrm{Q^{AB/F}}$&SSIM\\
    \midrule
    w/o intensity loss& 40.89& 0.85& 0.57&0.96\\
    w/o gradient loss& \textbf{43.77}& 0.86& 0.58&0.92\\
    w/o perception loss& 40.97& 0.75& 0.51&0.84\\
 w/o dynamic refresh& 42.82& 0.74& 0.55&0.84\\
    w/o distillation & 39.10& 0.86& 0.58&0.96\\  
 w/o digestible distillation& 36.14& 0.76& 0.53&0.89\\
 \rowcolor{gray!10}MMDRFuse& 43.08& \textbf{0.90}& \textbf{0.60}&\textbf{0.97}\\
\bottomrule
\end{tabular}
\vspace{-3mm}
\end{table}
\begin{table}
  \caption{Quantitative ablation about student model size of distillation on MSRS. The boldface shows the best value.}
  \captionsetup{skip=2pt}
  \label{tab:models size}
  \begin{tabular}{ccccc}
    \toprule
    Configurations&SD&VIF & $\mathrm{Q^{AB/F}}$&SSIM\\
\bottomrule
 huge student (1021.01KB)& 43.48& \textbf{0.97}& \textbf{0.63}&\textbf{0.99}\\
 large student (0.66KB)& \textbf{43.55}& 0.89& 0.59&0.93\\
 small student (0.22KB)& 38.35& 0.77& 0.50&0.91\\
 \rowcolor{gray!10}MMDRFuse (0.44KB)& 43.08& 0.90& 0.60&0.97\\
\bottomrule
\end{tabular}
\vspace{-4mm}
\end{table}
\subsubsection{Distillation Mechanism}
We design a student model with only two convolutional layers and 113 trainable parameters by utilising a powerful teacher.
Specifically, distillation is performed at the feature and response ends. 
To verify the effectiveness of distillation, we conduct an ablation study including the student model without distillation, the student model without digestible distillation, and our student model with digestible distillation.
From the last three rows of Table \ref{tab:ablation}, we can observe that the student can not learn effective information from the teacher without digestible distillation, resulting in a decrease in four metrics, which even worse than the student model without distillation. 
In contrast, through our digestible distillation, all four metrics achieve notable improvements. 
Positive evidence of these four metrics means the quality of the fused image itself is higher than before, sharing more detailed information with source images from both pixel level and visual perception perspectives. 

There are many volumes of targeted students available for us to choose from. 
To investigate the student sizes, we gradually compress the student model down to the minimum. 
As displayed in Table \ref{tab:models size},  from the perspective of metrics value, we should opt for the first student model (1021.01KB). 
However, in pursuit of an extremely mini model size, we choose to compress the model size to below 1KB. 
The model of 0.44KB in size is the optimal choice. 
 

Finally, we obtain the mini-model with only 113 trainable parameters. 
Before we display the fusion performance of our experiments, we first compare the model efficiency and complexity with several SOTA approaches, as displayed in Table \ref{model detail}. 
Among all the involved methods, our MMDRFuse exhibits the smallest model size, the least number of floating-point operations, and the fastest average running time. 
In the following experiments, we will further demonstrate that our MMDRFuse not only enjoys advantages in efficiency and complexity but also leads to promising model performance.
\begin{table}[h]
\centering
\caption{Comparison of Model Efficiency and Complexity with SOTA approaches. The boldface shows the best value.}
\captionsetup{skip=2pt}
\label{model detail}
\begin{tabular}{ccccc}
\toprule
 Methods&Venue& Size(KB)& Flops(G)& Time(MS)\\
\midrule
 TextFusion
&24' Arxiv& 288.51& 88.708& 11.17\\
 MUFusion
& 23' Inf. Fus.& 4333.76& 240.669&7.91\\
 LRRNet
&23' TPAMI& 192.20& 60.445& 523.35\\
 MetaFusion
& 23' CVPR& 3170.76& 1063.000&325.37\\
 DeFusion
& 22' ECCV& 30759.66& 322.696&339.59\\
 TarDAL
& 22' CVPR& 1158.50& 388.854&124.10\\

\rowcolor{gray!10}MMDRFuse& Ours& \textbf{0.44}& \textbf{0.142}&\textbf{0.62}\\
  \bottomrule
\end{tabular}
\vspace{-3mm}
\end{table}
\subsection{Infrared and Visible Image Fusion}
\subsubsection{Qualitative Comparison}
\begin{figure}[t]
\centering
\begin{minipage}[b]{\linewidth}
\centering
  \includegraphics[clip,width=0.7\linewidth]{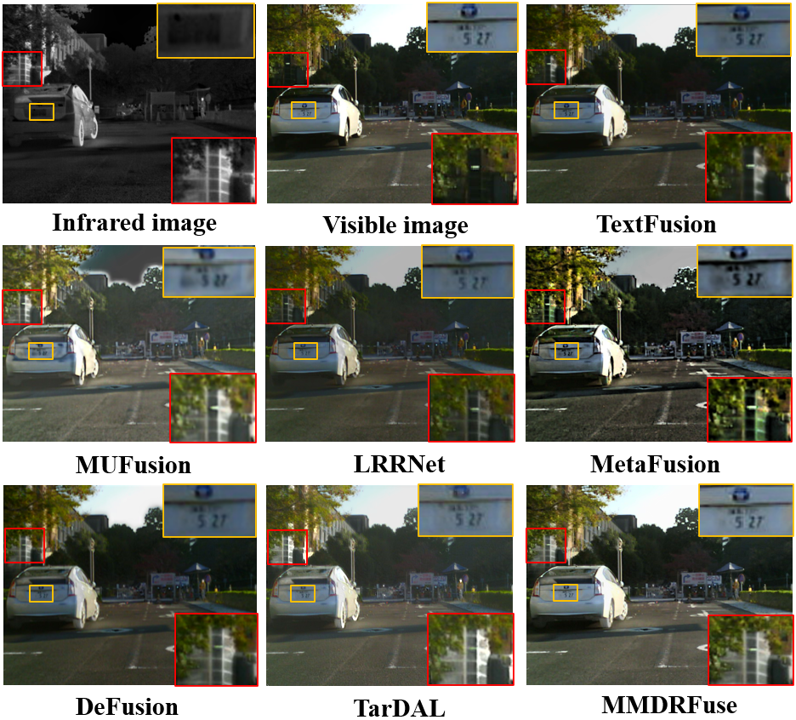}
  \vspace{-3mm}
  \caption{Visual comparison with SOTA approaches on MSRS.}
  \Description{Qualitative comparison of 9 methods on MSRS dataset.}
  \label{MSRS}
\end{minipage}
\hfill
\begin{minipage}[b]{\linewidth}
\centering
  \includegraphics[clip,width=0.7\linewidth]{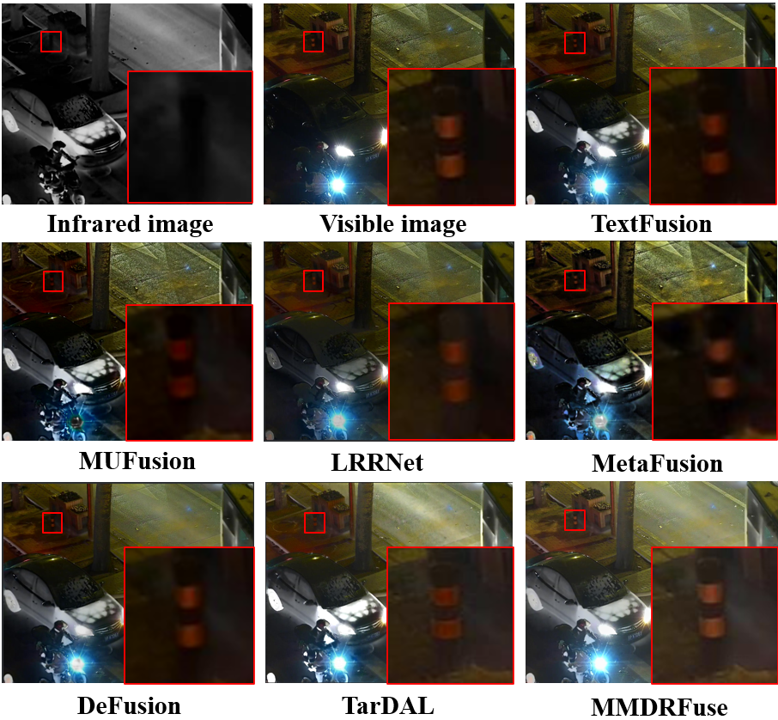}
  \vspace{-3mm}
  \caption{Visual comparison with SOTA approaches on LLVIP.}
  \Description{Qualitative comparison of 9 methods on LLVIP dataset.}
  \label{LLVIP}
\end{minipage}
\hfill
\begin{minipage}[b]{\linewidth}
\centering
  \includegraphics[clip,width=0.7\linewidth,height=5.5cm]{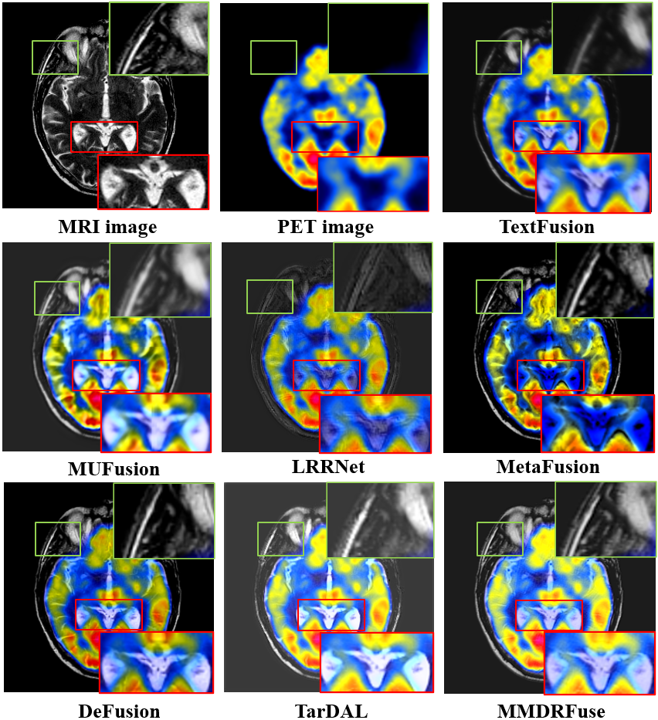}
  \vspace{-3mm}
  \caption{Visual comparison with SOTA approaches on MRI-PET.}
  \Description{Qualitative comparison of 9 methods on MRI-PET dataset.}
  \label{medical}
\end{minipage}
\end{figure}

We present the qualitative comparison results in Figure \ref{MSRS} and Figure \ref{LLVIP}.  
In Figure \ref{MSRS}, MUFusion, LRRNet, and MetaFusion not only distort the colour of the sky but also exhibit numerous artifacts around the license plate, as shown within the yellow box. 
Besides, DeFusion and TextFusion result in a blurry appearance of the leaves within the red box. 
TarDAL shows a relatively apparent contrast in the red box, but the colour of the building seems so bright that we can not clearly see the content within it. 
Our MMDRFuse can produce a clear result in the above points. 
Furthermore, our method can illuminate objects in the dark, even though those objects are not very clear in the source images, such as the pillar in Figure \ref{LLVIP}. 

\subsubsection{Quantitative Comparison}
To objectively demonstrate the advantage of MMDRFuse, we conduct quantitative experiments on three typical datasets, which include MSRS, LLVIP, and RoadScene. 
The MSRS dataset covers scenes from urban streets to rural roads under various lighting conditions, containing rich semantic information. 
The LLVIP dataset mainly includes urban and street environments under low-light conditions. 
The RoadScene dataset, on the other hand, contains scenes with roads, vehicles, pedestrians, and more under various lighting conditions. 

We present the quantitative results in Table \ref{IVIF}. 
Our method consistently ranks first in three metrics (SCD, $\mathrm{Q^{AB/F}}$, SSIM) across all the datasets.
Consistent performance can be obtained in VIF and CC, ranking first or second.
Besides, in terms of the SD metric that is susceptible to noise, we rank first on MSRS and third on RoadScene.

The outstanding performance in full-reference (SCD, VIF, $\mathrm{Q^{AB/F}}$, SSIM,CC) and no-reference (SD) indicators show that our fusion results not only closely resemble the source images at the visual and pixel levels, but also possess rich, high-quality information within each fused image itself. 
The above results also demonstrate that our mini model can adeptly handle a variety of scenes and lighting conditions, and it surpasses much larger models in both visual effects and evaluation metrics, perfectly achieving a balance between fusion performance and computation costs.

\begin{table}[h]
\centering
\captionsetup{skip=2pt}

\caption{Quantitative results of the IVIF task on different datasets. Boldface and underline show the best and second-best values, respectively.}
\label{IVIF}

\begin{subtable}{0.45\textwidth} 
\captionsetup{skip=2pt}
\caption{Dataset: MSRS Infrared-Visible Dataset }
\begin{tabular}{ccccccc}
\toprule
Methods&SD& SCD& VIF & $\mathrm{Q^{AB/F}}$ & SSIM &CC\\
\midrule
TextFusion&38.02& 1.43& 0.72& \underline{0.52}& 0.76 &0.59\\
MUFusion&28.48& 1.26& 0.60& 0.42& 0.71 &\underline{0.61}\\
LRRNet&31.76& 0.79& 0.54& 0.45& 0.43 &0.51\\
MetaFusion&\underline{39.64}& 1.50& 0.71& 0.48& 0.78 &0.60\\
DeFusion&34.86& 1.29& \underline{0.75}& 0.51& \underline{0.93} &0.60\\
TarDAL&37.58& \underline{1.53}& 0.70& 0.42& 0.70 &\textbf{0.63}\\
\rowcolor{gray!10}MMDRFuse&\textbf{43.08}& \textbf{1.66}& \textbf{0.90}& \textbf{0.60}& \textbf{0.97}&\underline{0.61}\\
\bottomrule
\end{tabular}
\end{subtable}
\hfill 
\begin{subtable}{0.45\textwidth} 
\captionsetup{skip=2pt}
\caption{Dataset: LLVIP Infrared-Visible Dataset }

\begin{tabular}{ccccccc}
\toprule
Methods&SD& SCD& VIF & $\mathrm{Q^{AB/F}}$& SSIM &CC\\
\midrule
TextFusion&47.82& 1.41& 0.71& \underline{0.53}& 0.77 &0.68\\
MUFusion&40.44& 1.03& 0.68& 0.47& 0.70 &0.65\\
LRRNet&35.42& 1.09& 0.56& 0.47& 0.65 &\underline{0.69}\\
MetaFusion&\underline{48.90}& 1.36& 0.61& 0.29& 0.60 &0.67\\
DeFusion&43.93& 1.23& \underline{0.74}& 0.43& \underline{0.83} &0.67\\
TarDAL&\textbf{61.77}& \underline{1.44}& 0.65& 0.42& 0.71 &0.66\\
\rowcolor{gray!10}MMDRFuse&47.74& \textbf{1.55}& \textbf{0.83}& \textbf{0.60}& \textbf{0.84} &\textbf{0.71}\\
\bottomrule
\end{tabular}
\end{subtable}
\begin{subtable}{0.45\textwidth} 
\textbf{\captionsetup{skip=2pt}}
\centering
\caption{Dataset: RoadScene Infrared-Visible Dataset}
\begin{tabular}{ccccccc}
\toprule
Methods&SD& SCD& VIF & $\mathrm{Q^{AB/F}}$& SSIM &CC\\
\midrule
TextFusion&38.43& 1.63& \textbf{0.68}& \textbf{0.44}& \underline{0.94} &0.49\\
MUFusion&\textbf{54.92}& 1.62& 0.51& 0.34& 0.78 &0.48\\
LRRNet&43.90& \underline{1.69}& 0.51& 0.38& 0.71 &\underline{0.51}\\
MetaFusion&\underline{48.66}& 1.60& 0.51& 0.36& 0.75 &\underline{0.51}\\
DeFusion&32.86& 1.48& 0.50& 0.38& 0.86 &0.48\\
TarDAL&44.88& 1.50& 0.53& \underline{0.39}& 0.83 &\textbf{0.52}\\
\rowcolor{gray!10}MMDRFuse&45.02& \textbf{1.70}& \underline{0.62}& \textbf{0.44}& \textbf{0.95} &\textbf{0.52}\\
\bottomrule
\end{tabular}
\end{subtable}
\vspace{-3mm}
\end{table}

\subsection{Medical Image Fusion}
\subsubsection{Qualitative Comparison}
To verify the generalisation ability of our method, we directly apply the above six SOTA methods and our design to the MIF task without retraining. 
We show the qualitative results of MRI-PET images in Figure \ref{medical}. 
From the green and red boxes, we can observe that our MMDRFuse not only preserves the detailed internal structural information displayed in MRI images but also retains the distribution of the radioactive tracers shown in PET images, providing a more comprehensive set of diagnostic information. 
While TextFusion,  LRRNet, and MetaFusion lose the information of internal structure. 
Besides, MUFusion, LRRNet, MetaFusion, and DeFusion dilute the distribution of the radioactive tracers. 
In terms of image clarity, MUFusion and TarDAL introduce extra artifacts in the detailed internal structure.

\subsubsection{Quantitative Comparison}
We display the quantitative results on both MRI-SPECT and MRI-PET image pairs in Table \ref{MIF}. 
Our MMDRFuse ranks first or second in five metrics (SD, SCD, VIF, $\mathrm{Q^{AB/F}}$, CC) and ranks third in one metric (SSIM) across the two datasets. 
Combined with the above qualitative results we can notice that TarDAL introduce extra artifacts that can reach high values of CC, while MetaFusion and DeFusion, diluting the distribution of the radioactive tracers, can obtain high values of SSIM, which can not provide much supportive information for diagnosis.
Therefore, after performing a comprehensive analysis of the above results, our MMDRFuse with an extremely tiny mode size and considerable performance is better suited to meet the requirements of medical image fusion.
\begin{table}[h]
\captionsetup{skip=2pt}

\caption{Quantitative results of the IVIF task. Boldface and underline show the best and second-best values, respectively.}
\label{MIF}
\begin{subtable}{0.45\textwidth} 
\caption{Dataset: MRI-SPECT Infrared-Visible Dataset }
\begin{tabular}{ccccccc}
\toprule
Methods&SD& SCD& VIF & $\mathrm{Q^{AB/F}}$& SSIM    &CC\\
\midrule
TextFusion&41.32& 0.31& 0.52& 0.21& 0.32 &\underline{0.87}\\
MUFusion&54.19& 0.91& 0.46& 0.44& 0.36 &0.85\\
LRRNet&42.16& 0.42& 0.34& 0.20& 0.21 &0.84\\
MetaFusion&45.43& 0.73& 0.47& 0.39& \underline{1.40} &0.86\\
DeFusion&51.33& 0.70& \textbf{0.59}& \underline{0.55}& \textbf{1.47} &0.86\\
TarDAL&\underline{58.41}& \underline{1.19}& \underline{0.56}& 0.48& 0.37 &0.78\\
\rowcolor{gray!10}MMDRFuse&\textbf{65.67}& \textbf{1.78}& \textbf{0.59}& \textbf{0.57}& 0.39 &\textbf{0.88}\\
\bottomrule
\end{tabular}
\end{subtable}

\begin{subtable}{0.45\textwidth} 
\caption{Dataset: MRI-PET Infrared-Visible Dataset }
\captionsetup{skip=2pt}
\begin{tabular}{ccccccc}
\toprule
Methods&SD& SCD& VIF & $\mathrm{Q^{AB/F}}$& SSIM    &CC\\
\midrule
TextFusion&60.1& 1.29& \textbf{0.63}& 0.36& 0.38 &0.79\\
MUFusion&60.81& 1.22& 0.43& 0.42& 0.38 &0.77\\
LRRNet&48.30& 0.55& 0.37& 0.21& 0.22 &0.73\\
MetaFusion&\underline{63.47}& 1.26& 0.50& \underline{0.51}& \textbf{1.43} &0.77\\
DeFusion&49.26& 1.11& 0.45& 0.45& \underline{1.18} &0.64\\
TarDAL&52.99& 1.1& 0.46& 0.45& 0.30 &\textbf{0.84}\\
\rowcolor{gray!10}MMDRFuse&\textbf{71.50}& \textbf{1.72}& \underline{0.60}& \textbf{0.56}&  0.43&\underline{0.81}\\
\bottomrule
\end{tabular}
\end{subtable}
\vspace{-3mm}
\end{table}
\subsection{Pedestrian Detection}
\begin{figure}[t]
\centering
\includegraphics[clip,width=0.9\linewidth]{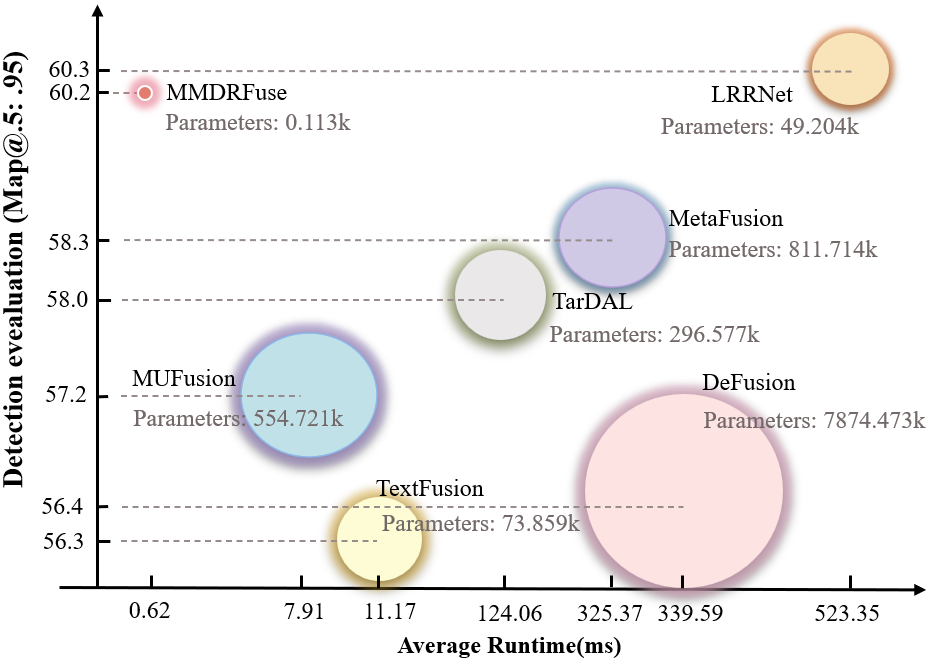}

\caption{Comparison in terms of average detection precision and average speed for RGB-T pedestrian detection.
}
\Description{Comparison of Pedestrian Detection Performance .}
\label{detection}
\vspace{-3mm}
\end{figure}
To further validate the performance of our MMDRFuse on downstream detection tasks, we conduct experimental exploration on pedestrian detection. 
As can be seen from Figure \ref{detection}, although LRRNet~\cite{li2023lrrnet} achieves the best detection precision, it requires more than $\times$850 times of computation consumption compared to our design. 
Furthermore, compared with MetaFuse~\cite{MetaFusion}, which introduces the detection task as additional supervision for fusion, our method performs better even without any prior settings in detecting pedestrians. 
This can be attributed to the introduction of perceptual loss, combined with the dynamic refresh strategy and distillation mechanism that jointly learns advanced semantic information.
Our method can precisely detect pedestrians within contrasting backgrounds.
Our fusion model enable supporting RGB-T detection without involving semantic information as input, with less running time as well as a mini model size under 1 KB, achieving satisfactory detection results at a considerably marginal cost. 

\section{Conclusion}
An investigation into compressing the image fusion model is conducted in this work.
We utilise the specially designed comprehensive loss function and the dynamic refresh strategy based on intermediate fusion results to first formulate a fusion model.
Further combined with the digestible distillation strategy, we successfully train an extremely tiny (0.44 KB) student network from the teacher model.
Experimental results on multiple tasks demonstrate that our mini model not only exhibits advantages in efficiency and complexity but also achieves promising results on the MMIF and downstream detection tasks.

\begin{acks}
This work was supported in part by the National Natural Science Foundation of China (62106089, 62020106012, 62332008, 62336004), and in part by the Engineering and Physical Sciences Research Council (EPSRC), U.K. (Grants  EP/V002856/1, and EP/T022205/1).
\end{acks}

\bibliographystyle{ACM-Reference-Format}
\balance
\bibliography{sample-base}


\end{document}